\documentclass[10pt]{article} 
\usepackage[preprint]{tmlr}

\usepackage{amsmath,amsfonts,bm}

\def\eqref#1{equation~\ref{#1}}

\def\1{\bm{1}}

\DeclareMathAlphabet{\mathsfit}{\encodingdefault}{\sfdefault}{m}{sl}
\SetMathAlphabet{\mathsfit}{bold}{\encodingdefault}{\sfdefault}{bx}{n}

\usepackage[dvipsnames]{xcolor}
\usepackage{hyperref}
\usepackage{url}
\usepackage{graphicx}
\usepackage{booktabs}
\usepackage{multirow}
\usepackage{float}
\usepackage{listings}
\usepackage{adjustbox}
\usepackage{subcaption}
\usepackage{cleveref}
\title{On the modality gap and the contrastive loss in multi-modal representation learning}

\author{\name Fabian Mager \email fmager@dtu.dk \\
      \addr  Department of Applied Mathematics and Computer Science\\
      Technical University of Denmark
      \AND
      \name Hiba Nassar \email hibna@dtu.dk \\
      \addr Department of Applied Mathematics and Computer Science\\
      Technical University of Denmark
      \AND
      \name Lars Kai Hansen \email lkai@dtu.dk \\
      \addr Department of Applied Mathematics and Computer Science\\
      Technical University of Denmark
}

\def\month{MM}  
\def\year{YYYY} 
\def\openreview{\url{https://openreview.net/forum?id=XXXX}} 

\usepackage{xspace}

\newcommand\rmix{r\textsubscript{mix}\xspace}
\renewcommand\u{\mathbf{u}}
\renewcommand\v{\mathbf{v}}
\newcommand\A{\mathbf{A}}
\newcommand\U{\mathcal{U}}
\newcommand\V{\mathcal{V}}
\newcommand{\InfoNCE}{InfoNCE\xspace}
\newcommand{\XinfoNCE}{xNCE\xspace}
\renewcommand\L{\mathcal{L}}
\newcommand{\ssize}{\scriptsize}
\newcommand{\spm}[1]{{\ssize$\pm$#1}}
\newcommand\ul[1]{\underline{#1}}
\newtheorem{hyp}{Hypothesis}
\newcommand{\Star}[1]{#1\ensuremath{^\star}\kern-\scriptspace}
\newcommand{\Diam}[1]{#1\ensuremath{^\diamond}\kern-\scriptspace}
\newcommand{\Trian}[1]{#1\ensuremath{^\triangleright}\kern-\scriptspace}
\newcommand{\ua}{{\ssize$\uparrow$}}
\newcommand{\da}{{\ssize$\downarrow$}}

\newcommand{\eg}{e.g.\@\xspace}
\newcommand{\ie}{i.e.\@\xspace}

\begin{document}

\maketitle

\begin{abstract}
	We study the \emph{modality gap} in CLIP-style dual-encoder contrastive learning, where image and text embeddings remain misaligned despite being trained in a shared space. We argue that the gap is induced by a failure of the \InfoNCE formulation with independent encoders. We conduct a uni-modal experiment with two independent encoders and identical initialization conditions and find that \InfoNCE \emph{actively} generates a gap at low temperatures. We provide a theoretical analysis of this phenomenon and show that the modality gap is indeed a mode-failure of \InfoNCE, but only at low temperatures. We propose a simple modification called \XinfoNCE, which uses intermodal as well as intra-modality negative contrastive pairs. \XinfoNCE matches retrieval performance on MS-COCO while consistently reducing the gap even at low temperatures. Notably, \XinfoNCE improves zero-shot classification over the \InfoNCE baseline across all benchmarks, whereas high-temperature \InfoNCE and regularized \InfoNCE both fail to do so, demonstrating that \XinfoNCE reduces the modality gap without sacrificing the discriminative geometry needed for transfer. Code availability: \href{}{...}
\end{abstract}

\section{Introduction}
\label{sec:intro}

Contrastive learning has emerged as a central paradigm for self-supervised representation learning. \citet{WU_2018} introduced the view-invariant instance discrimination objective that underpins modern contrastive methods. Subsequent advances such as \emph{SimCLR} \citep{CHEN_2020b} demonstrated that strong data augmentation, large batch sizes, and a contrastive loss are sufficient to produce high‑quality visual representations, while \citet{BACHMAN_2019} broadened the theoretical foundation by defining contrastive learning as maximizing mutual information between augmentations. Parallel developments, including \emph{CPC} \citep{OORD_2019} and \emph{MoCo} \citep{HE_}, further refined contrastive objectives through predictive coding and momentum-based negative sampling. Building on these foundations, \emph{CLIP} (Contrastive Language Image Pretraining) extended contrastive learning to the multimodal domain \citep{RADFORD_2021}.\\
\emph{Multimodal learning} aims to build representations that jointly model information from heterogeneous data sources (\eg, images and language), enabling systems to exploit complementary cues and reason about shared semantics across modalities \citep{RADFORD_2021}. They furthermore showed that pairing images with natural‑language descriptions in a contrastive learning paradigm yields highly transferable and zero‑shot‑capable models at scale.\\
CLIP uses one encoder per modality, trained using a contrastive loss function called Information Noise-Contrastive Estimation (\InfoNCE). For a batch of images and captions, \eg, the loss aims to predict the correct caption for a given image and vice versa, based on pairwise cosine similarities. The authors show that such a pretraining objective produces embeddings with superior zero-shot performance compared to purely vision-based self-supervised models \citep{RADFORD_2021}.\\
Despite this success, CLIP-style dual-encoder models often exhibit a pronounced \emph{modality gap}: image and text embeddings, while comparable via cosine similarity, tend to occupy distinct regions (or submanifolds) of the joint representation space rather than forming a fully mixed distribution. Previous analyses show that this separation has geometric and optimization roots, including inductive biases in initialization and training dynamics in contrastive learning that maintain a distance between modalities, influenced by factors such as temperature, initialization, dimensionality, and data heterogeneity \citep{LIANG_a, SHI_2023, FAHIM_2024d, ROLE_2025, HUANG_}.

\section{Related Work}
\label{sec:related_work}

Recent work has examined the \emph{modality gap} in CLIP, i.e., the tendency of image and text embeddings to occupy distinct, non‑overlapping regions of the shared representation space. Several explanations have been proposed in the literature.
\cite{LIANG_a} attribute the gap to modality-specific cone structures present at random initialization and optimization dynamics. They find that embeddings of ResNets, image encoders, and text transformer architectures form narrow cones on the unit sphere, which are highly separable across random initializations. The contrastive loss fails to close this gap during optimization. Under constrained optimization conditions (\ie mismatched pairs), they find that when shifting the modalities toward each other, the contrastive loss increases, suggesting that the modality gap configuration is a local minimum of the contrastive loss at low temperatures. This local minimum disappears as the temperature increases or if the mismatched pairs are removed. \citet{SHI_2023} identify conflicting properties in the \InfoNCE loss function. They argue based on proof-of-concept experiments on the 3D unit sphere that the gap corresponds to local minima of the CLIP objective, making it difficult for optimization to align the modalities. \citet{FAHIM_2024d} experimentally show that the modality gap in CLIP is neither a result of data heterogeneity nor initialization conditions. They experiment with \InfoNCE in low dimensions and report that the modality gap does not arise. To circumvent the problem of the modality gap in high dimensions, they add regularization to the standard \InfoNCE loss, which promotes close proximity of positive pairs as well as a uniform distribution of embeddings on the unit sphere, inspired by \citet{WANG_2020}. The regularization reduces the gap and improves zero-shot performance on image classification benchmarks. Other work also shows that mitigating the gap can improve downstream performance. \citet{ROLE_2025} provide principled metrics showing that reducing the gap can improve retrieval and zero‑shot classification, while \citet{HUANG_} showed that preserving or compensating for the gap can benefit continuous learning. \citet{ESLAMI_2024} proposes regularizing the contrastive loss using an intra-modal separation term, which again shows better retrieval and downstream utility. \citet{OH_2023} propose a new finetuning method, which defines hard negative samples by mixing modalities, resulting in more uniform embeddings. Other research points to deeper geometric and algorithmic factors: \citet{LEVI_2025} find that image and text embeddings lie on distinct ellipsoidal shells encoding modality-dependent uncertainty. Collectively, these findings describe the modality gap as an inherent geometric and optimization feature of contrastive models.
While some of these studies attribute the modality gap to properties of the contrastive objective, a mechanistic understanding of its emergence is still lacking. Current improvements to \InfoNCE attempt to \emph{mitigate} the gap by introducing additional loss terms with opposing effects (\eg \citep{FAHIM_2024d, GRASSUCCI_2025, ESLAMI_2024}), rather than fixing the divergent properties of \InfoNCE itself.

\section{Background}
\label{sec:background}

\subsection{Contrastive learning via \InfoNCE}
\label{sec:infonce}

Let $\left[x_i, y_i\right]_{i=1}^N$ be a minibatch of paired samples from two modalities.
We denote the encoders by $f: \mathcal{X} \to \U \in \mathbb{R}^{N \times D}$ and $g: \mathcal{Y} \to \V \in \mathbb{R}^{N \times D}$, and write
$\u_i = f(x_i)$ and $\v_j = g(y_j)$. We assume $(x_i, y_i)$ to be positive pairs and $(x_i, y_k), \, k\neq i$ to be negative pairs. Let $a_{ij} \in\; [-1, 1]$ be the cosine similarity between $\u_i$ and $\v_j$, where $a_{ii}$ and $a_{ik}$ are similarity scores for positive and negative pairs, respectively. For a fixed anchor $x_i$ (i.e., fixed $\mathbf{u}_i$) and its targets $\{\mathbf{v}_j\}_{j=1}^N$, the softmax probability $P_i(\tau)$ with temperature $\tau > 0$ is

\begin{equation}
	P_i(\tau)=\frac{\exp\!\big(a_{ii}/\tau\big)}{\sum_{j=1}^{N} \exp\!\big(a_{ij}/\tau\big)}\;.
	\label{eq:softmax}
\end{equation}

The \InfoNCE loss from $\U \to \V$ is the mean over the negative $\log$-probabilities $P_i(\tau)$:

\begin{equation}
	\L_{\U\to \V}(\tau) \;=\; \mathbb{E}\left[- \log P_{i}(\tau)\right]\;.
	\label{eq:InfoNCE}
\end{equation}

The equivalent loss term $\L_{\V \to \U}$ is obtained by switching the indices $i$ and $j$ in \cref{eq:softmax} and \cref{eq:InfoNCE}, and the final loss value is the average of the two. The loss decreases as we increase the similarity of positive pairs $a_{ii}$ and decrease the similarities of negative pairs $a_{ik}$. Typically, all embeddings are $\ell_2$-normalized, and similarity is computed using the dot product. \citet{WANG_2020} investigate the properties of representations learned via a contrastive loss, and propose two quantities: \emph{alignment} (closeness of features from positive pairs) and \emph{uniformity} (of the distribution of features on the hypersphere). Alignment is achieved by maximizing the similarity of positive pairs, and uniformity is achieved by maximizing the entropy of the feature distribution on the hypersphere. They show that for a \emph{uni-modal} setup, \ie where $\u$ and $\v$ are processed by the same encoder, these properties are asymptotically optimized, and correlate strongly with downstream performance.

\subsection{The Temperature Parameter}

The temperature $\tau$ scales the logits in the softmax and thereby controls the sharpness of the distribution $P_i(\tau)$. Smaller values of $\tau$ amplify small similarity gaps, resulting in a winner-takes-all probability distribution, while larger values of $\tau$ attenuate those gaps and produce a more uniform probability distribution. Common values for $\tau$ are in the range $[0.01, 1]$, with CLIP having a learnable temperature parameter, which is initialized as $0.07$ and converging to $0.01$ during training \citep{RADFORD_2021}.

\subsection{Mode-Failure due to Dual-Encoders}

In the uni-modal scenario, $\U$ and $\V$ are outputs of a single parametrized function. Updating the parameters of this function will affect both $\U$ and $\V$. The only way to accurately classify a positive pair from negative pairs is to learn discriminative features of $\u$ and $\v$, which are invariant to noise factors (e.g. augmentations). The contrastive loss asymptotically optimizes for alignment of positive pairs and uniformity of the feature distribution on the hypersphere \citep{WANG_2020}.
In a multi-modal scenario, $\U$ and $\V$ are outputs of independently parametrized functions. Although the parameters are optimized under a joint objective, the parameter sets do not share weights. Thus, while both encoders may learn equivalently good discriminative features, it is not guaranteed that positives are mapped to the same region on the hypersphere, as they are encoded independently. While the numerator in \cref{eq:softmax} encourages alignment of positives, the denominator can be satisfied by moving $\u$ and $\v$ far from each other, providing a trivial objective and explaining the phenomenon of the modality gap. Thus, we hypothesize that the uniformity term in the multi-modal \InfoNCE objective is no longer only an information preservation term on $\U$ and $\V$, but also a degenerative term with a divergence objective on $\U$ and $\V$. We believe that by eliminating this degenerative divergence objective, we can eliminate the modality gap and improve the performance of the model.

This motivates the following two hypotheses.
\begin{hyp}[H1]
	\label{hyp:1}
	\emph{Induced gap by multimodal \InfoNCE formulation}.
	The modality gap is the result of an \emph{active inductive bias} from the \InfoNCE formulation with independent encoders and the induced optimization dynamics, rather than being a byproduct of data idiosyncrasies, initialization, dimensionality or temperature.
\end{hyp}
\begin{hyp}[H2]
	\label{hyp:2}
	\emph{Mitigating the gap by sampling intra- and inter-modality pairs}.
	The modality gap can be reduced by sampling negative pairs $a_k$ interchangeably from both modalities, which removes the divergence objective of two modalities. We call this adaptation \emph{``Mixed Noise-Contrastive Estimation''}, or \emph{\XinfoNCE}.
\end{hyp}

\section{Mixing modalities - \XinfoNCE}
\label{sec:xinfonce}

\begin{figure}[t]
	\centering
	\includegraphics[width=.95\linewidth]{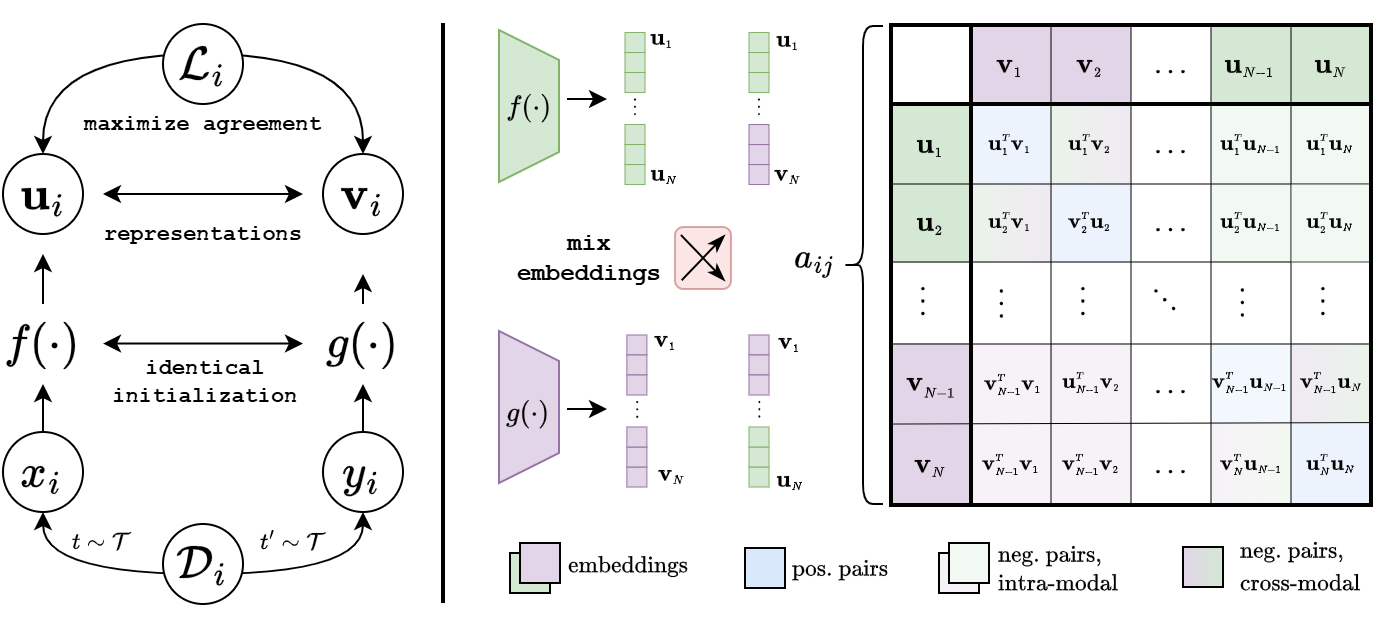}
	\caption{\textbf{Left}: A \emph{SimCLR} like investigation of the multimodal \InfoNCE loss. We generate two views of the same datapoint from the same family of augmentations $t \sim T$ and $t' \sim T$, and pass the two views to identically initialized, independent encoders $f(\cdot)$ and $g(\cdot)$ (\ie no weight sharing). Adapted from \cite{CHEN_2020b} \textbf{Right}: Our proposed method \XinfoNCE to avoid modality shifts. Rather than only sampling cross-modal negative pairs, we include intra- and cross-modality negative pairs. Adapted from \cite{RADFORD_2021}.}
	\label{fig:method}
\end{figure}

\begin{figure}[t]
	\begin{minipage}{.4\textwidth}
		\centering
		\begin{minipage}{.95\textwidth}
			\begin{lstlisting}[language=Python]
# Inputs:
# I   - Image emb. [n, d_i]
# T   - Text emb. [n, d_t]
# r_mix   - mixing ratio, float
# temp  - temperature param.
# ce_loss - Cross Entropy Loss

# mix modalities
bs = I_e.size(0)    # batch size
mask = np.arange(bs) < int(bs*r_mix)
I_mix = np.where(mask, T, I)
T_mix = np.where(mask, I, T)

# scaled pairwise cos. sim.
logits = np.dot(I_mix, T_mix.T)
logits /= temp

# symmetric loss function
labels = np.arange(n)
loss_i = ce_loss(logits, labels)
loss_t = ce_loss(logits.T, labels)
loss = (loss_i + loss_j) / 2
            \end{lstlisting}
		\end{minipage}
	\end{minipage}
	\begin{minipage}[c]{.6\textwidth}
		\hfill
		\includegraphics[width=0.9\textwidth]{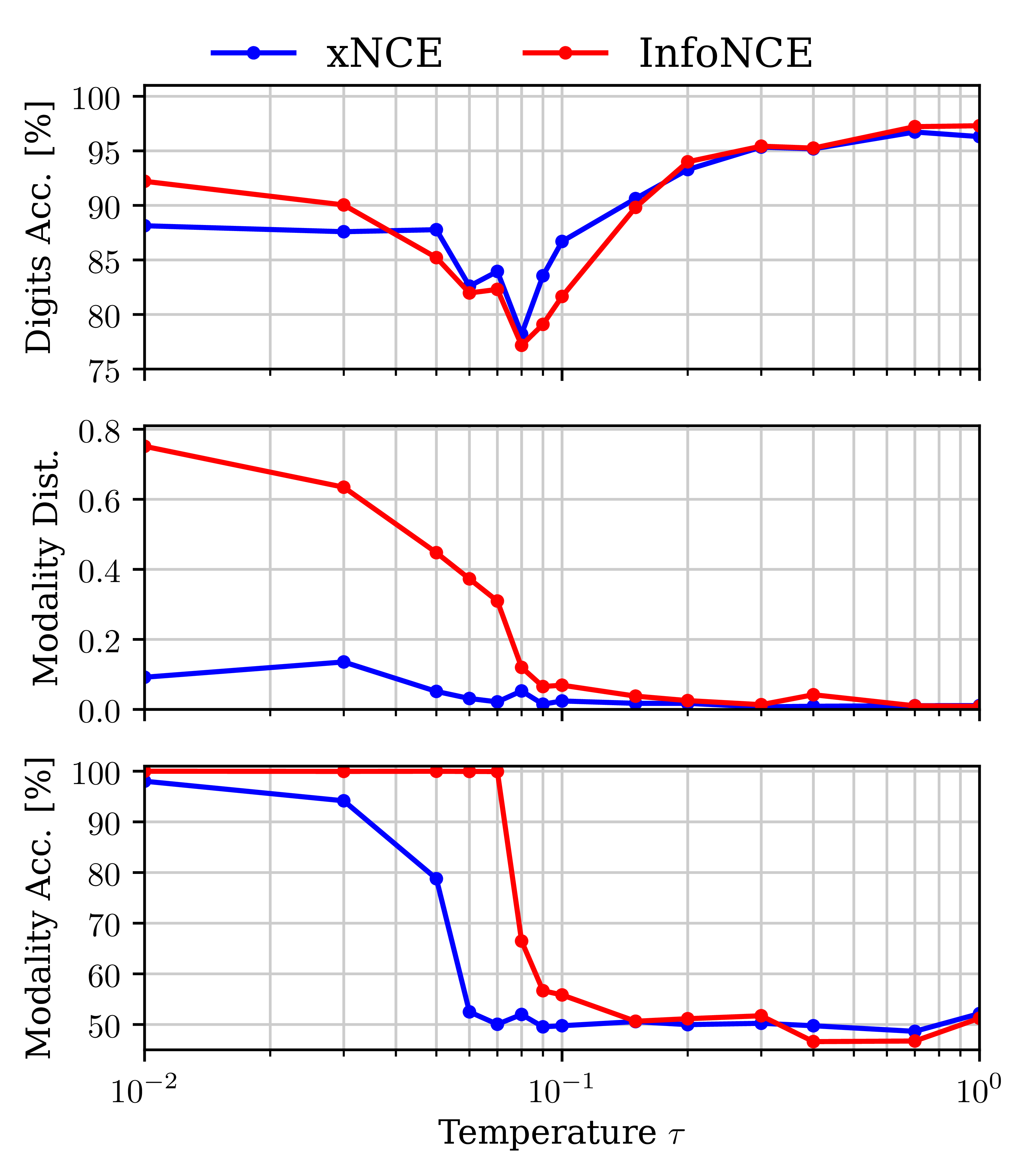}
	\end{minipage}
	\caption{\textbf{Left}: The proposed \XinfoNCE loss function. Numpy like pseudo-code for mixing image and text embeddings. \textbf{Right}: Model behaviour for different temperature values. We measure classification accuracy of a linear probe, modality distance as euclidean distance of modality centroids and modality classification accuracy using a linear probe.}
	\label{fig:code}
\end{figure}

We propose a simple and efficient modification to the \InfoNCE formulation of \citet{RADFORD_2021}. Our method builds on the following assumption for an ideal multimodal latent space: Two modalities are fully aligned and occupy the same space if the modality type of a latent variable becomes unidentifiable. If so, we can use the latent variables from both modalities interchangeably in the loss formulation in \cref{eq:InfoNCE}. For a given anchor $\u_i$, there exists only one positive $\v_i$. However, we can use any $\u_k$ or $\v_k$, $k \neq i$, as a negative pair. We refer to negative pairs from the same modality, \ie $\u_i$ and $\u_k$, as \emph{intramodality} negatives and pairs from different modalities, \ie $\u_i$ and $\v_k$, as \emph{intermodality} pairs. The procedure is depicted in \cref{fig:method} (right) and a numpy like pseudo-code is shown in \cref{fig:code}. When calculating the pairwise cosine similarity $a_{ij}$ for mixed embeddings, the diagonal will remain the same. The off-diagonal elements will consist of $N\times r_{mix}$ intra-modality pairs and $N \times (1-r_{mix})$ inter-modality pairs, where \rmix is the mixing ratio applied to $\U$ and $\V$. Because the final loss is a symmetric cross entropy loss over the similarity scores, the mixing is also symmetric around $r_{mix}=0.5$.
Most importantly, this formulation removes the degenerate divergence objective of \InfoNCE. In \XinfoNCE, the denominator of \cref{eq:InfoNCE} includes intermodality negatives, leading to an increase in loss as the modalities drift apart.

\subsection{Unimodal: Image-to-Image using MNIST}
\label{sec:mnist_experiments}

We hypothesize that the modality gap is the result of the multi-modal \InfoNCE loss formulation, rather than initialization and data idiosyncrasies (Hyp. \ref{hyp:1}). We therefore set up a \emph{SimCLR} \citep{CHEN_2020b} like experiment, where two views are generated from the same source using the same family of augmentations \cref{fig:method} (left). In contrast to \citet{CHEN_2020b}, we use two independent encoders, \ie without weight sharing or momentum updates, as is the case in a multi-modal setup. We use the same architecture and number of parameters for both encoders. Importantly, we initialize both encoders with the same weights, which ensures identical embeddings at initialization for the same input. If a gap is observed, it would indicate that the gap is not a passive byproduct but an active element in the optimization dynamics of \InfoNCE. A short overview of this setup is given in \cref{fig:method}.

\paragraph{Dataset.} We use the \emph{digit} split of the Extended MNIST (EMNIST) dataset \citep{EMNIST}, which contains ten handwritten digit classes. To construct contrastive pairs, we generate two independently augmented views of the same image. Training augmentations include random rotation ($\pm 45^{\circ}$), random resized cropping (scale $\in [0.7, 1.0]$), color jitter, and additive Gaussian noise with standard deviation $\sigma = 0.3$. Both views are normalized with the EMNIST channel statistics ($\mu{=}0.1307$, $\sigma{=}0.3081$) and passed to its respective encoder. Thus, 'modality' in the context of the Image-to-Image contrastive learning refers to the output of each encoder.

\paragraph{Model.} We use a minimalistic vision transformer \citep{DOSOVITSKY_2020} tailored for $28 \times 28$ grayscale images, with patch size $4$, embedding dimension $d{=}48$, $4$ attention heads, $4$ transformer layers, and an MLP ratio of $4$. The transformer employs pre-norm residual blocks with GELU activations and 2D sinusoidal positional encodings. A learnable \texttt{[CLS]} token is prepended to the patch sequence, and its final-layer representation serves as the image embedding. Embeddings are $\ell_2$-normalized before computing the contrastive loss.

\paragraph{Loss.} We train several models with the \InfoNCE and \XinfoNCE objective. We use a fixed mixing ratio of $r_{mix}{=}0.5$ for all MNIST experiments. We keep the temperature $\tau$ a fixed parameter and iterate on in total 14 values, ranging from $0.01$ to $1.0$.

\subsection{Multimodal: Image-to-Text using COCO}
\label{sec:coco_experiments}

For a truly multimodal task, we use the Microsoft Common Object in Context Dataset (COCO) \citep{MSCOCO} and finetune CLIP \citep{RADFORD_2021}. COCO is widely used for benchmarking vision language models, and is not included in the pretraining corpus of CLIP. All results presented in this paper are on the Kaparthy split test set \citep{KAPARTHY}, obtained from \citet{DELPHBOY}. COCO has five captions per image. For training, we choose a random caption among the five, whereas for validation and testing, we use the first caption.

\paragraph{Dataset.} The COCO 2017 dataset \citep{MSCOCO}, which pairs images with natural language captions. Each image is associated with multiple captions; during training, a caption is sampled uniformly at random per image, while at validation time the first caption is used deterministically. Images are preprocessed using the standard CLIP preprocessing pipeline \citep{RADFORD_2021}. Captions are tokenized using CLIP's byte-pair encoding tokenizer. Object category annotations from the COCO instances set provide a multi-label classification signal over 80 object categories, encoded as multi-hot vectors.

\paragraph{Model.} We finetune both vision and text encoders. We use the CLIP ViT-B/16 obtained from \cite{CLIP_GH}.

\paragraph{Loss.} \InfoNCE serves as the baseline. As CLIP converges to a temperature of $0.01$, we set this to be the initial value of $\tau$. We used both a learnable temperature parameter where we clip the value of $\tau$ if $<0.01$, equivalent to \citet{RADFORD_2021}, as well as a cosine annealed temperature parameter with $\tau_{max}=0.7$. For all base experiments, we do not use a contrastive head \citep{OORD_2019} as part of the loss function. The loss is calculated on the output of the encoders. We additionally compare against regularized InfoNCE (Reg) from \cite{FAHIM_2024d}, which augments the InfoNCE objective with explicit alignment, cross-modal uniformity, and unimodal uniformity terms, inspired by the work of \citet{WANG_2020}. For implementation details, see section \ref{sec:appendix_regularization}. For \XinfoNCE, we do not know the optimal temperature value. We tried several initial temperature values in the range of 0.01 to 0.05 and found that $\tau{=}0.02$ worked marginally better. We trained with three different mixing ratios: \rmix=0.01, 0.05, and 0.5.

\subsection{Evaluation}
\label{sec:evaluation}

To assess the structure of the learned representations, we train linear probes evaluating (i) \emph{the separability} of the modality and (ii) \emph{the downstream utility}. All probes are trained concurrently with the main contrastive objective on the training data. Crucially, all probes serve purely as diagnostic tools without influencing representation learning. For (i), we train a single linear layer using a binary cross entropy loss to predict the modality. As for (ii), we train the probes per modality and report the average score. For MNIST experiments, we use the digit class as target (single-label, C=10) and report classification accuracy. For COCO, we use the instances represented in the image (multi-label, C=80), available through the object-detection task in COCO and report the multilabel average precision. In addition, we measure the \emph{alignment} and \emph{uniformity} properties proposed by \citet{WANG_2020}. Alignment is defined as the expected $\ell_2$-distance between positive pairs, and uniformity is defined as the expected pairwise Gaussian potential, measured for each modality separately. For Image-to-Text experiments, we evaluate the image-to-text (I2T) and text-to-image (T2I) retrieval recalls at ranks 1, 5, and 10 (R@1, R@5 and R@10, respectively). To evaluate the transferability of the learned image representations, we also conduct zero-shot classification on four image-only benchmarks: CIFAR-10 (N=10k, C=10), CIFAR-100 (N=10k, C=100), Caltech-101 (N$\approx$9.7k, C=100), and ImageNet (N=50k, C=1000). For each dataset, we follow \citet{RADFORD_2021} and encode class names using the prompt template \emph{"A photo of a \{class\}"} and classify images by nearest neighbour in embedding space under cosine similarity. We report top-1 and top-5 accuracy. No fine-tuning or adaptation is performed; the encoder weights are used as-is after COCO training.

\subsection{Optimization}
\label{sec:optimization}

All models are trained using AdamW with a weight decay of 0.04 and a three-phase learning rate schedule: \emph{Phase 1 - probes only}: Near-zero learning rate, allowing the linear probes to warm up on randomly initialized representations before the encoders begin learning. Probe optimizers use a fixed learning rate throughout training. \emph{Phase 2 - warmup}: Linear warmup to the base learning rate. \emph{Phase 3 - decay}: Cosine annealing to $1e-2\times$ the base learning rate. We used a single NVIDIA Ampere GPU for the Image-to-Image experiments and a single NVIDIA H100 GPU for the Image-to-Text experiments.
All key hyperparameters for both Image-to-Image (\cref{sec:mnist_experiments}) and Image-to-Text experiments (\cref{sec:coco_experiments}) are listed in \cref{tab:model_params,tab:opt_params} in \cref{sec:appendix_hyperparams}.

\begin{figure}[t]
	\centering
	\includegraphics[width=.8\linewidth]{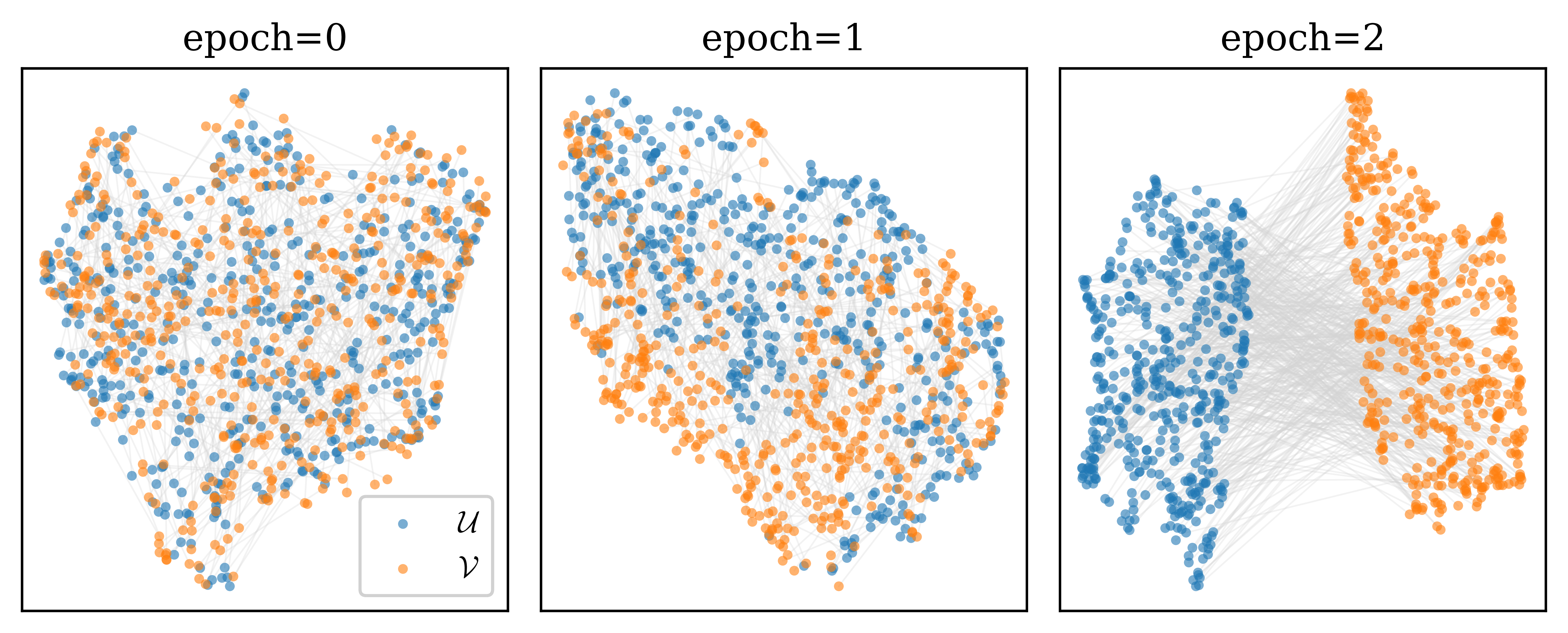}
	\caption{Two dimensional UMAP projection of MNIST embeddings for training using \InfoNCE at low temperature ($\tau{=}0.01$). Points are coloured by their respective encoder. Embeddings overlap at initialization as both encoders are initialized with identical weights, and get separated immediately after.}
	\label{fig:mnist_umap}
\end{figure}

\section{Results}

\subsection{Unimodal: Image-to-Image using MNIST}
\label{sec:mnist_results}

We investigate how mixing modalities in the \InfoNCE loss formulation, such that negative similarities include both intra- and inter-modality pairs, affects the embedding space. We name this adaptation \XinfoNCE. We measure digit and modality classification accuracy via a linear probe, as well as the modality centroid distance. \cref{fig:code} compares these measures for \InfoNCE and \XinfoNCE over a range of temperature values.

\paragraph{Digit classification accuracy.} We find a lower classification accuracy for a temperature of $ 0.03 < \tau < 0.15$, and a higher classification accuracy towards low and high temperatures $\tau< 0.03$ and $\tau > 0.2$. The best digit separability of $97.2\%$ and $96.7\%$  is achieved at $\tau{=}1.0$ and $\tau{=}0.7$ for \InfoNCE and \XinfoNCE, respectively. The margin of \InfoNCE and \XinfoNCE is slightly larger at lower temperatures, where classification accuracy is $92.2\%$ and $88.1\%$ for $\tau{=}0.01$, respectively.

\paragraph{Modality Distance.} For \InfoNCE, we find that the modality distance is $0.75$ at $\tau{=}0.01$, monotonically decreasing to a value of $0.07$ at $\tau{=}0.09$, and approximately constant for $\tau \ge 0.09$. Compared to \InfoNCE, the modality gap for \XinfoNCE is generally less for all values of $\tau$, starting at $0.09$ and $0.13$ for $\tau{=}0.01$ and $\tau{=}0.03$ and converging to less than $0.05$ for $\tau{\ge}0.05$.

\paragraph{Modality classification accuracy.}
The modality accuracy for \InfoNCE is at $100\%$ for low temperatures and ${\approx}50\%$ at high temperatures, with a steep transition at $\tau \approx 0.09$. The \XinfoNCE shows high classification accuracies for lower temperatures but slightly below $100\%$ and $\approx 50\%$ for $\tau \ge 0.07$, with a more subtle transition.

\begin{figure}[t]
	\centering
	\begin{minipage}{\textwidth}
		\begin{minipage}{0.49\textwidth}
			\includegraphics[width=1.0\linewidth]{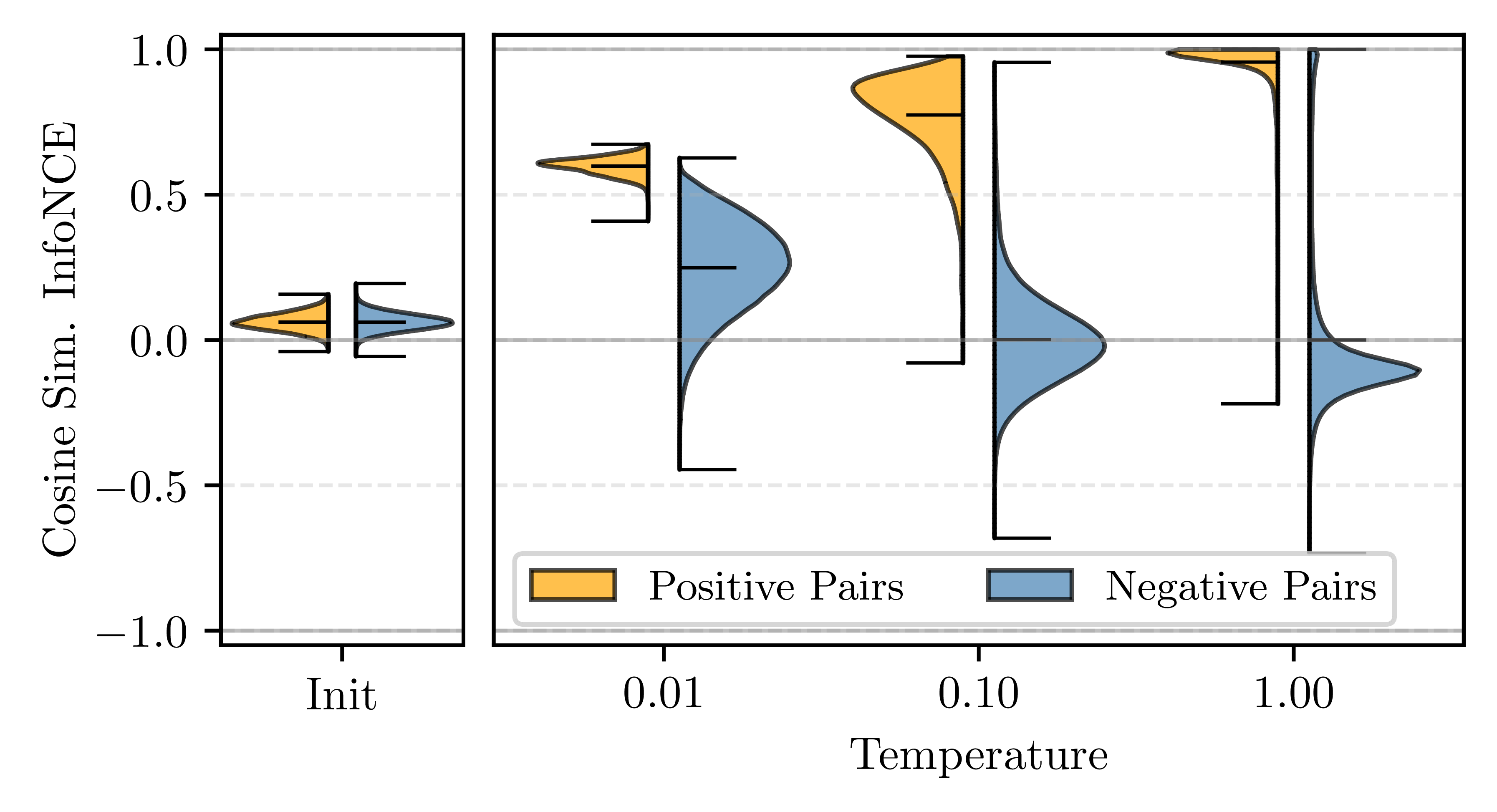}
		\end{minipage}
		\begin{minipage}{0.49\textwidth}
			\includegraphics[width=1.0\linewidth]{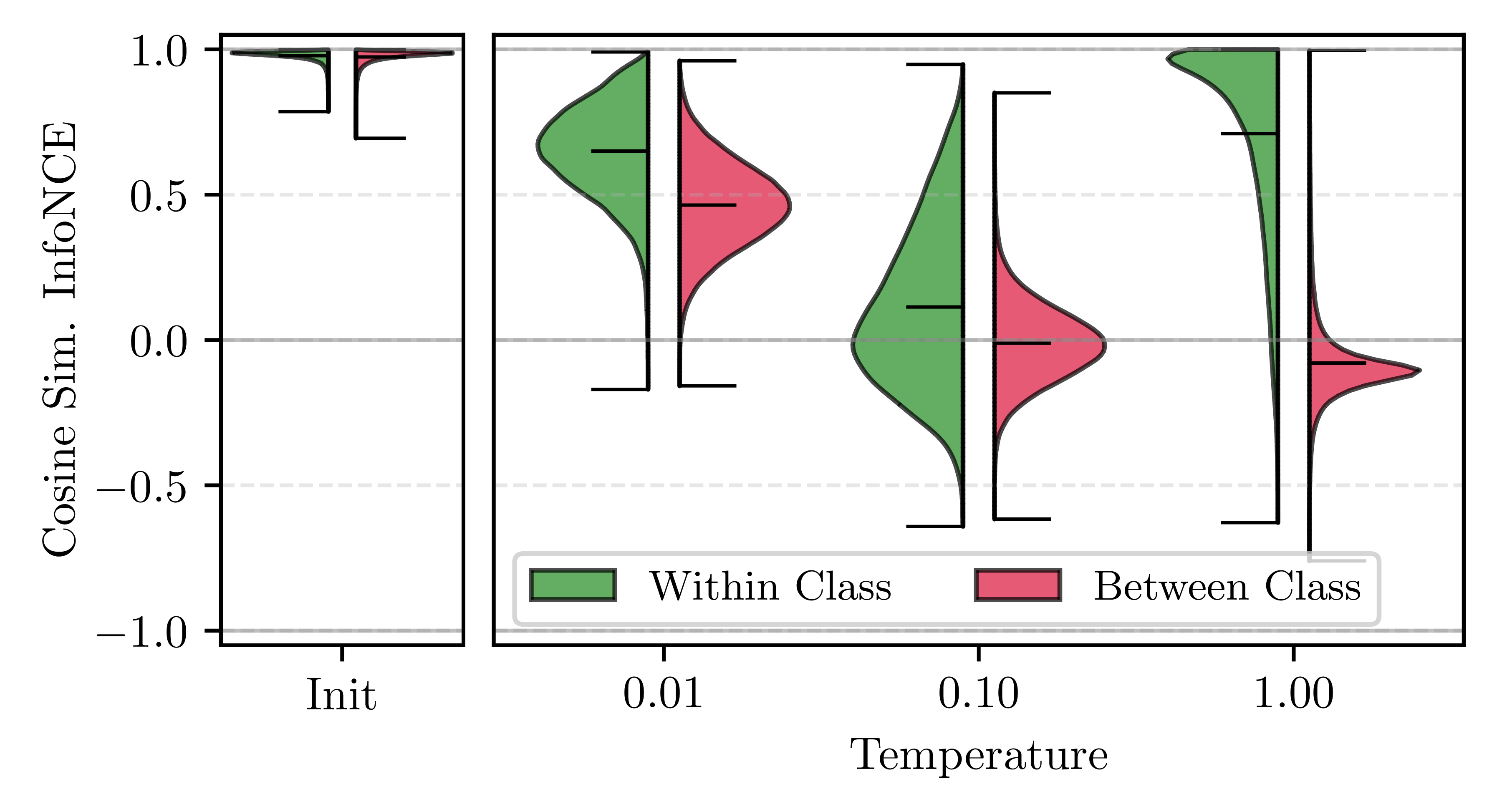}
		\end{minipage}
	\end{minipage}
	\begin{minipage}{\textwidth}
		\begin{minipage}{0.49\textwidth}
			\includegraphics[width=1.0\linewidth]{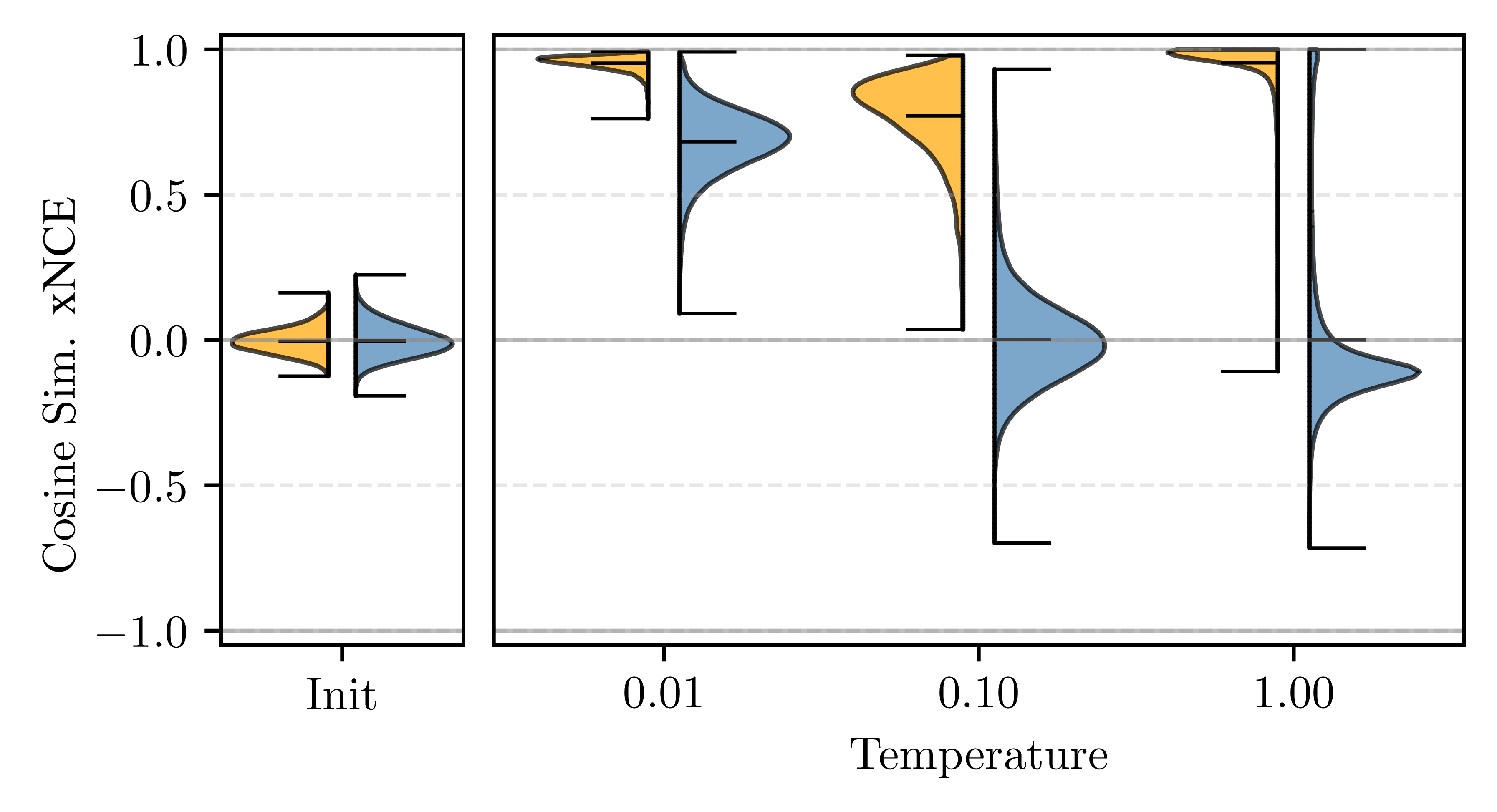}
		\end{minipage}
		\begin{minipage}{0.49\textwidth}
			\includegraphics[width=1.0\linewidth]{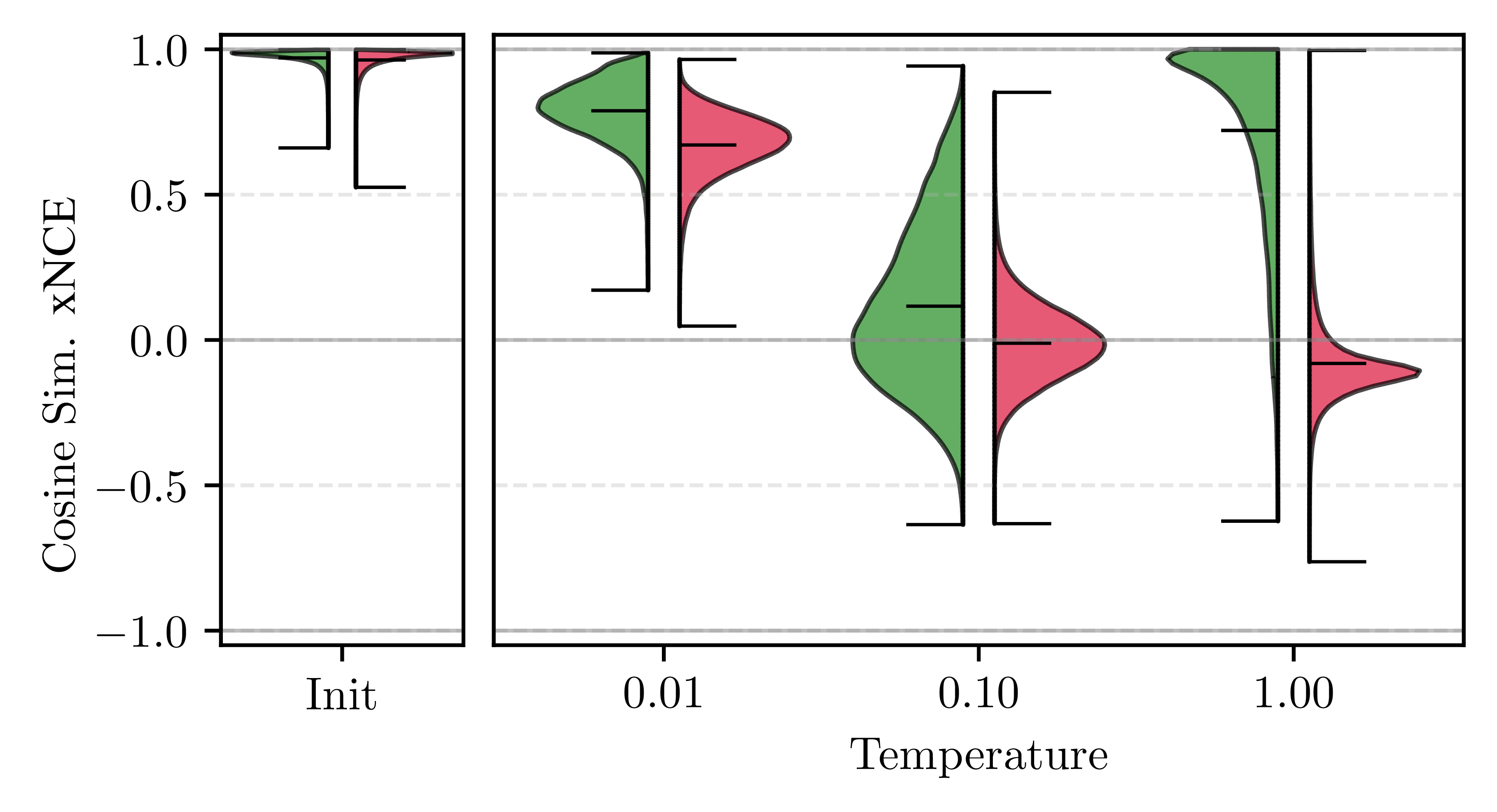}
		\end{minipage}
	\end{minipage}
	\caption{Distribution of similarity scores over different temperatures for the original \InfoNCE loss (\textbf{top}) and our \XinfoNCE loss (\textbf{bottom}). \textbf{Left}: Cosine similarity scores for positives pairs ($a_{ii}$) and negatives pairs ($a_{ik}$). Low temperatures show short tails in the distribution of negative pairs, whereas higher temperatures show lower variance but longer tails. \XinfoNCE shows higher scores for in $\tau{=}0.01$, but behaves similarly for $\tau{=}0.1$ and $\tau{=}1.0$. \textbf{Right}: Cosine similarity scores conditioned on class labels. The distributions are furthest apart for $\tau{=}1.0$. No or very little difference between \InfoNCE and \XinfoNCE.}
	\label{fig:simdistribution}
\end{figure}

\paragraph{Latent geometry under different temperatures.} Figure \cref{fig:simdistribution} (left) shows the cosine similarity scores at different temperatures $\tau$. For \InfoNCE, low temperatures result in distributions with positive similarities $a_{ii} \ll 1$ and negative similarities $a_{ik} \gg 0$. For higher temperatures, the distribution is stretched and slightly skewed, where positive similarities are close to $1$ and negative similarities range from $-1$ to $1$ with a mean of $\approx 0$. The distribution of pairwise similarities for low temperatures indicates that $\V$ is located in a narrow cone far from the query $\u_i$, which aligns with the observed modality gap. When using \XinfoNCE, the modality gap disappears, as the positive similarities are very close to $1$. Mixing modalities has little or no effect on the similarity distribution for higher temperatures. Temperature also affects the within-modality distribution (\cref{fig:simdistribution}, right). Although the linear separability of classes for low and high temperatures is similar (\cref{fig:code}, right), the classes occupy a much narrower space for low temperatures (\eg $\tau{=}0.01$) than they do for high temperatures (\eg $\tau{=}1.0$).

\subsection{Theoretical Analysis of the Temperature in InfoNCE}

We analyze the limit $\lim_{\tau \to 0^+} \L_i$ (\Cref{eq:InfoNCE}), as this is where we observe the modality gap. First, we establish a formal definition of the phenomenon. As \Cref{eq:InfoNCE} acts on the similarity scores $a_{ij}$, we define the modality gap as follows:

Let $\mathcal{M}: \mathcal{X}\times\mathcal{Y}\to\U\times\V$ denote a parametrized function, which maps the inputs $x$ and $y$ to their corresponding latent variables $\u$ and $\v$. For pairwise cosine similarities $\A \in \mathbb{R}^{N\times N}$ with elements $a_{ij} \in [-1, 1]$, we denote positive similarities of $\u_i$ and $\v_i$ as $a_{ii}$ and the negative cosine similarities of $\u_i$ and $\v_k$, where $k \neq i$ as $a_{ik}$.


\paragraph{Universal approximator}. For perfectly unified and aligned spaces $\U$ and $\V$, we have $a_{ii} = 1$ for all $i$, and $a_{ik} < a_{ii}$ for all $k \neq i$. The expected pairwise similarity $\mathbb{E}[a_{ik}] = 0$, as this is consistent with $\U$ and $\V$ being uniformly distributed on the hypersphere. Also note that $\L \to 0$ for $\tau \to 0$, as $\lim_{\tau \to 0} P_i(\tau)=1$.

\paragraph{Constrained approximator}. If we assume an irreducible approximation error, the cost for the condition where $P\!\left(a_{ii} \leq \max_{k\neq i} a_{ik}\right) > 0$ depends on the margin $\max_{k\neq i} a_{ik} - a_{ii}$ and the temperature $\tau$. For smaller values of $\tau$, the cost for misclassified samples increases. Note that higher values of $\tau$ conversely decrease the gain of correctly classified samples. The role of $\tau$ on the local geometry is a well studied phenomenon \citep{MANNA_2025, WANG_2021a}.

We can rewrite \cref{eq:InfoNCE} as the sum of the negative \emph{energy} term ${a_{ii}}/{\tau}$ and the \emph{entropy} term $\log \left(\sum_{j=1}^N e^{a_{ij} / \tau}\right)$. We define $m_i = \frac{1}{N}\sum_{j=1}^N a_{ij}$ and write

\begin{equation}
	\L_i(\tau)=-\log P_{i}(\tau)=-\frac{a_{ii}-m_i}{\tau}+\log \left(\sum_{j=1}^N e^{(a_{ij}-m_i) / \tau}\right)\,.
	\label{eq:energy}
\end{equation}

%




%
The modality gap arises for small temperatures:

\begin{equation}
	\lim_{\tau \to 0^+}\L_i(\tau) \approx -\frac{a_{ii}-m}{\tau} + \frac{\max(a_{ij}) - m}{\tau} \,.
	\label{eq:lowtau}
\end{equation}

If we assume an irreducible approximation error, \ie the optimal ranking $[a_{ij}]_{i,j}^N$ with $a_{ii} \neq \max(a_{ij})$, we have $\L_i \approx \left(\max(a_{ij}) - a_{ii}\right)/{\tau}$. The cost can then be further reduced without changing the ranking of $[a_{ij}]$ by letting all $a_{ij}$ converge toward their mean $m_i$, reducing the pairwise margins and resulting in a degenerate optimum of \cref{eq:lowtau}.

\paragraph{The modality gap phenomenon.} For a given vector $\u_i$, its positive $\v_i$ and negatives $\v_k$ for all $k \neq i$, a narrow similarity distribution $[a_{ij}]$ where $a_{ik} \approx m_i$ while fixing the ranking of $[a_{ij}]$ corresponds to compressing all $\v_k$ into a tight cluster. If we add ${a_{ii}} \to 1$ and $a_{ik} \to 0$ for $\tau > 0$, we have $0 < m_i < 1$, which implies that all $\v_k$ form a tight cluster distant from $\u_i$, which gives us the definition of the modality gap.




\begin{figure}[tb]
	\centering
	\includegraphics[width=.85\textwidth]{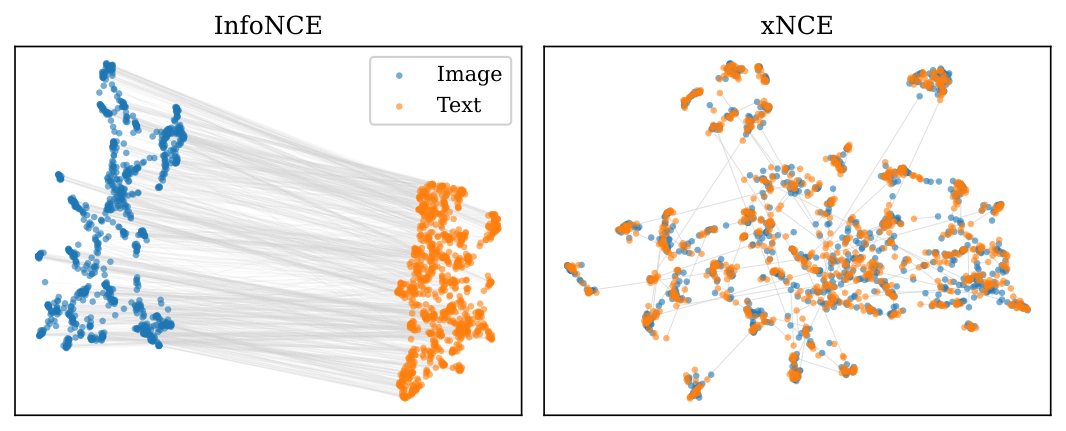}
	\caption{Visualization of Image (\textcolor{MidnightBlue}{blue}) and Text (\textcolor{RedOrange}{orange}) embeddings of COCO after finetuning with \InfoNCE (left) and \XinfoNCE (right) using UMAP \citep{UMAP}. Matching pairs are connected with grey lines. \InfoNCE shows a clear separation of modalities, whereas \XinfoNCE does not.}
	\label{fig:umap}
\end{figure}

\subsection{Multimodal: Image-to-Text using COCO}
\label{sec:res-img2txt}

\begin{table}[bt]
	\centering
	\caption{COCO metrics using \InfoNCE (Org), regularized \InfoNCE (Reg), and \XinfoNCE (Mix) with different mixing ratios \rmix. Overall best performing model indicated using bold font, best performance \XinfoNCE (Mix) models are underlined.}
	\begin{adjustbox}{max width=\textwidth}
		\begin{tabular}{lcccccccc}
			\toprule
			\multirow{2}[3]{*}{Loss} & \multirow{2}[3]{*}{$\tau$} & \multirow{2}[3]{*}{\rmix} & \multicolumn{2}{c}{Class mAP} & \multicolumn{2}{c}{Modality} & \multicolumn{2}{c}{Embeddings}                                                                           \\
			\cmidrule(lr){4-5}\cmidrule(lr){6-7}\cmidrule(lr){8-9}
			                         &                            &                           & Image (\ua)                   & Text (\ua)                   & Dist. (\da)         & Acc. (\%, \ua)       & Align. (\da) & Unif. (\da) \\
			\midrule
			Org                      & \Star{.01}                 & -                         & 18.24\spm{3.25}               & 11.26\spm{0.82}              & 0.88\spm{0.02}      & 99.98\spm{0.02}      & 1.38         & 0.06        \\
			Org                      & \Diam{.70}                 & -                         & \textbf{70.45}\spm{0.04}          & \textbf{67.38}\spm{0.04}         & \textbf{0.02}\spm{0.00} & \textbf{50.12}\spm{0.12} & \textbf{0.16}    & 0.03        \\
			\midrule[0.1pt]
			Reg                      & \Star{.01}                 & -                         & 59.98\spm{0.43}               & 53.65\spm{0.50}              & 0.12\spm{0.00}      & 80.40\spm{0.15}      & 0.70         & \textbf{0.02}   \\
			\midrule[0.1pt]
			Mix                      & \Star{.02}                 & .01                       & 17.91\spm{0.60}               & 17.53\spm{0.09}              & 0.12\spm{0.01}      & 74.32\spm{2.45}      & 0.83         & \ul{0.04}   \\
			Mix                      & \Star{.02}                 & .05                       & 23.32\spm{1.67}               & 21.83\spm{1.67}              & 0.08\spm{0.00}      & 61.03\spm{1.31}      & 0.67         & 0.05        \\
			Mix                      & \Star{.02}                 & .50                       & \ul{28.89}\spm{3.96}          & \ul{27.82}\spm{3.34}         & \ul{0.05}\spm{0.00} & \ul{54.02}\spm{1.00} & \ul{0.48}    & 0.08        \\
			\bottomrule
			\\
			\addlinespace[-1ex]\multicolumn{9}{l}{\footnotesize{\Star{} are learnable parameter \hspace{2.5mm}\Diam{} fixed parameter \hspace{2.5mm}}}                                                                                 \\
			\label{tab:mscoco_emb_metrics}
		\end{tabular}
	\end{adjustbox}
\end{table}

\begin{table}[btb]
	\centering
	\caption{COCO retrieval performance for \InfoNCE (Org), regularized \InfoNCE (Reg), and \XinfoNCE (Mix), where R@K = Recall at rank K.}
	\begin{adjustbox}{max width=\textwidth}
		\begin{tabular}{lcccccccc}
			\toprule
			\multirow{2}[3]{*}{Loss} & \multirow{2}[3]{*}{$\tau$} & \multirow{2}[3]{*}{\rmix} & \multicolumn{3}{c}{Image to Text} & \multicolumn{3}{c}{Text to Image}                                                                                                  \\
			\cmidrule(lr){4-6}\cmidrule(lr){7-9}
			                         &                            &                           & R@1                               & R@5                               & R@10                 & R@1                  & R@5                  & R@10                      \\
			\midrule
			Org                      & \Star{.01}                 & -                         & 75.83\spm{0.18}                   & \textbf{96.51}\spm{0.05}              & \textbf{99.18}\spm{0.09} & \textbf{75.52}\spm{0.02} & \textbf{96.04}\spm{0.13} & \textbf{98.87}\spm{0.04}      \\
			Org                      & \Diam{.70}                 & -                         & 49.67\spm{0.05}                   & 83.67\spm{0.09}                   & 92.83\spm{0.04}      & 47.90\spm{0.02}      & 82.32\spm{0.10}      & 92.40\spm{0.08}           \\
			\midrule[0.1pt]
			Reg                      & \Star{.01}                 & -                         & \textbf{76.31}\spm{0.18}              & 96.37\spm{0.05}                   & 98.96\spm{0.00}      & 74.79\spm{0.14}      & 95.87\spm{0.10}      & 98.67\spm{0.04}           \\
			\midrule[0.1pt]
			Mix                      & \Star{.02}                 & .01                       & \ul{75.82}\spm{0.21}              & 96.41\spm{0.08}                   & \ul{99.11}\spm{0.08} & 75.48\spm{0.12}      & 95.81\spm{0.21}      & \ul{\textbf{98.87}}\spm{0.02} \\
			Mix                      & \Star{.02}                 & .05                       & 75.65\spm{0.10}                   & \ul{96.43}\spm{0.01}              & \ul{99.11}\spm{0.02} & \ul{75.51}\spm{0.06} & \ul{95.91}\spm{0.06} & 98.79\spm{0.05}           \\
			Mix                      & \Star{.02}                 & .50                       & 75.18\spm{0.10}                   & 96.12\spm{0.11}                   & 98.86\spm{0.04}      & 75.35\spm{0.28}      & 95.89\spm{0.08}      & 98.73\spm{0.05}           \\
			\bottomrule
			\\
			\addlinespace[-1ex]\multicolumn{9}{l}{\footnotesize{\Star{} are learnable parameter \hspace{2.5mm}\Diam{} fixed parameter}}                                                                                                                                \\
		\end{tabular}
	\end{adjustbox}
	\label{tab:mscoco_retrieval_metrics}
\end{table}

\paragraph{Latent-space structure and modality gap.} Image-to-Text contrasts using \InfoNCE (Org) and \XinfoNCE (Mix) on CLIP ViT-B/16 fine-tuned on COCO. With a learnable temperature, \InfoNCE exhibits a large modality gap (distance $0.88\pm{0.02}$; linear modality separability $99.98\pm{0.02}\%$) and a comparatively low instance semanticity in the embeddings (image mAP $18.24\pm{3.25}$; text mAP $11.26\pm{0.82}$). Regularized \InfoNCE (Reg) substantially reduces the gap (distance $0.15\pm{0.01}$; modality accuracy $82.51\pm{0.62}\%$) but at a higher computational cost of approximately 25--30\% longer training time compared to standard \InfoNCE and \XinfoNCE. In contrast, \XinfoNCE reduces the modality gap even at small mixing ratios (\eg $\rmix{=}0.01$: distance $0.12\pm{0.01}$; modality accuracy $74.32\pm{2.45}\%$) while improving instance mAP over the \InfoNCE baseline ($17.91\pm{0.60} \to 28.89\pm{3.96}$ for image; $11.26\pm{0.82} \to 27.82\pm{3.34}$ for text, as \rmix increases from 0.01 to 0.50). Increasing the mixing ratio further tightens the modalities (distance 0.12$\to$0.05; modality accuracy 74.32\%$\to$54.02\%) and continues to increase instance semanticity, revealing a controllable trade‑off between modality separability and semanticity. For reference, a scheduled temperature increase in \InfoNCE from $\tau{=}0.01$ to $\tau{=}0.70$ eliminates the modality gap (distance $0.02\pm{0.00}$; modality accuracy $50.12\pm{0.12}\%$) and yields the highest instance mAP (image $70.45\pm{0.04}$; text $67.38\pm{0.04}$). \Cref{fig:modality_dist_progression} in \cref{sec:appendix_convergence} shows the modality distance during training: \XinfoNCE immediately reduces the gap, whereas Reg shifts it more gradually, indicating that xNCE actively counteracts gap formation rather than regularizing it post-hoc.

\paragraph{Retrieval performance.} \Cref{tab:mscoco_retrieval_metrics} reports retrieval recall R@K for I2T and T2I. The \InfoNCE model with learnable temperature attains the strongest overall retrieval compared to \XinfoNCE by a relatively small margin. Regularized \InfoNCE (Reg) achieves the best I2T R@1 ($76.31\pm{0.18}$) while T2I R@1 ($74.79\pm{0.14}$) trails standard \InfoNCE slightly, offering better modality alignment overall. \XinfoNCE with small mixing retains near-parity at all K, with a slight decrease as \rmix increases. We provide a more detailed sensitivity analysis of \rmix on retrieval performance in \Cref{fig:sensitivity} in \Cref{sec:sensitivity}. The \InfoNCE with high temperatures yields the worst retrieval performance, dropping to $\approx 15\%$ at R@1 in I2T compared to the low-temperature setting.


\begin{table}[btb]
	\centering
	\caption{Zero-shot classification performance for \InfoNCE (Org), regularized \InfoNCE (Reg), and \XinfoNCE (Mix), reporting Top-1 and Top-5 accuracy (\%) on Caltech-101, CIFAR-10, CIFAR-100, and ImageNet}
	\begin{adjustbox}{max width=\textwidth}
		\begin{tabular}{lccccccccccc}
			\toprule
			\multirow{2}[3]{*}{Loss} & \multirow{2}[3]{*}{$\tau$} & \multirow{2}[3]{*}{\rmix} & \multicolumn{2}{c}{Caltech-101} & \multicolumn{2}{c}{CIFAR-10} & \multicolumn{2}{c}{CIFAR-100} & \multicolumn{2}{c}{ImageNet}                                                                                     \\
			\cmidrule(lr){4-5}\cmidrule(lr){6-7}\cmidrule(lr){8-9}\cmidrule(lr){10-11}
			                         &                            &                           & Top-1                           & Top-5                        & Top-1                         & Top-5                        & Top-1              & Top-5              & Top-1              & Top-5              \\
			\midrule
			Org                      & \Star{.01}                 & -                         & 82.4\spm{1.0}                   & 98.2\spm{0.0}                & 84.7\spm{0.4}                 & 99.1\spm{0.1}                & 56.9\spm{1.1}      & 82.6\spm{0.3}      & 51.9\spm{0.3}      & 79.2\spm{0.3}      \\
			Org                      & \Diam{.70}                 & -                         & 55.7\spm{0.4}                   & 84.0\spm{0.5}                & 80.2\spm{0.2}                 & 99.0\spm{0.0}                & 32.1\spm{0.2}      & 60.0\spm{0.2}      & 15.5\spm{0.1}      & 34.4\spm{0.2}      \\
			\midrule[0.1pt]
			Reg                      & \Star{.01}                 & -                         & 79.6\spm{0.2}                   & 96.5\spm{0.2}                & 84.2\spm{0.6}                 & 99.4\spm{0.1}                & 50.7\spm{0.6}      & 78.0\spm{0.6}      & 43.6\spm{0.2}      & 71.6\spm{0.2}      \\
			\midrule[0.1pt]
			Mix                      & \Star{.02}                 & .01                       & \textbf{83.1}\spm{0.3}              & \textbf{98.2}\spm{0.2}           & \textbf{87.6}\spm{0.3}            & 99.4\spm{0.1}                & \textbf{61.1}\spm{0.3} & \textbf{85.1}\spm{0.4} & \textbf{54.0}\spm{0.1} & \textbf{82.4}\spm{0.1} \\
			Mix                      & \Star{.02}                 & .05                       & 82.7\spm{0.4}                   & 98.2\spm{0.0}                & 86.1\spm{0.6}                 & 99.2\spm{0.1}                & 57.4\spm{0.3}      & 82.8\spm{0.6}      & 50.7\spm{0.1}      & 79.6\spm{0.1}      \\
			Mix                      & \Star{.02}                 & .50                       & 80.9\spm{0.4}                   & 98.1\spm{0.1}                & 86.3\spm{0.3}                 & \textbf{99.5}\spm{0.1}           & 55.2\spm{0.5}      & 81.8\spm{0.4}      & 46.3\spm{0.2}      & 75.6\spm{0.1}      \\
			\bottomrule
		\end{tabular}
	\end{adjustbox}
	\label{tab:zero_shot}
\end{table}

\paragraph{Zero-shot performance.}
\Cref{tab:zero_shot} reports zero-shot classification on Caltech-101, CIFAR-10, CIFAR-100, and ImageNet. \XinfoNCE with small mixing ($\rmix{=}0.01$) achieves the best Top-1 accuracy on all four benchmarks, outperforming the \InfoNCE baseline by $+0.7$, $+2.9$, $+4.2$, and $+2.1$ percentage points, respectively. Performance degrades monotonically as \rmix increases, mirroring the retrieval trade-off: larger mixing ratios reduce the modality gap but erode the discriminative structure needed for zero-shot transfer. Regularized \InfoNCE (Reg) consistently trails the \InfoNCE baseline despite its lower modality distance, suggesting that its alignment regularization comes at the cost of transferable visual features. The high-temperature \InfoNCE variant ($\tau{=}0.70$) reduces zero-shot performance sharply, most severely on ImageNet (Top-1 $51.9{\to }15.5\%$), confirming that the geometry favored by low temperatures improves discriminative power.

\section{Discussion}
\paragraph{Temperature and mixing control representation geometry.}
Our empirical and theoretical analyses point to a coherent picture: \InfoNCE for independent encoders has a mode-failure at low temperatures $\tau$, which governs the similarity distribution and, in turn, the geometry of the learned space (\cref{sec:evaluation,sec:coco_experiments}). Training independent encoders on digits using \InfoNCE yields perfectly separable modalities at low temperatures (modality accuracy = 100\%; distance $\approx$ 0.75 at $\tau{=}0.01$) and a sharp transition towards modality overlap near $\tau{\approx}$ 0.09. In our theoretical analysis, we show that low temperatures produce a more narrow similarity distribution, as the cost of false positives can be reduced by shrinking similarity scores towards their mean. For larger $\tau$, the distribution has more flexibility in the tails, as more points share the weight for a false positive. This aligns well with the modality gap phenomenon, as a decreasing $\tau$ with an increasingly narrow distribution eventually restricts the range of positive similarities with an upper limit smaller than 1.

\paragraph{Mixing modalities prevents the modality gap.} Introducing \XinfoNCE moderates this effect by adding intra- and intermodal pairs to the negative sets. Mixing samples from both modalities to create negative pairs effectively removes the mode-failure, as simple shrinkage of similarity scores between modalities would change the ranking of points, resulting in a large penalty. For Image-to-Image experiments, \XinfoNCE reduces the modality gap uniformly across $\tau$ (distance ${\le}0.13$ already at $\tau{\le}0.03$ and distance ${<}0.05$ for $\tau{\ge}0.05$) and softens the transition in modality separability. In Image-to-Text experiments, \XinfoNCE reduces the gap compared to \InfoNCE with a learnable temperature (distance $0.88{\to}0.12$ at \rmix=0.01; further to 0.05 at \rmix=0.50) and simultaneously increases instance-level semanticity (image/text mAP increasing steadily with \rmix), indicating that mixed negatives act as an implicit regularizer that promotes cross-modal overlap without erasing within-modality class structure.
The results reveal a tunable tradeoff: (i) modality separability, (ii) within‑modality semanticity (\eg, linear‑probe mAP/accuracy), and (iii) retrieval precision. In Image-to-Text, \XinfoNCE with small mixing (\rmix $\in$ {0.01,0.05}) substantially reduces the gap while maintaining near‑parity with \InfoNCE in R@5/10 and trailing only slightly at R@1.
In contrast, increasing $\tau$ aggressively within \InfoNCE also reduces the modality gap and increases the linear‑probe mAP but significantly degrades retrieval (\eg $\approx$ 15 percentage points at R@1 versus low‑temperature \InfoNCE). Although both temperature and mixing influence the same underlying geometry, mixing offers a more favorable training method for retrieval: small \rmix materially lowers the gap and increases semanticity with negligible loss at R@5/10 and only minor impact at R@1, especially when paired with a projection head. It is also worth mentioning that we finetune a model using \XinfoNCE on top of pretrained weights using \InfoNCE, which is not favoring our method.

\paragraph{Mixing modalities improves zero-shot performance.}
The zero-shot results reinforce the picture established by the retrieval and embedding metrics, but with an important additional finding: \XinfoNCE with small mixing ($\rmix{=}0.01$) improves zero-shot Top-1 accuracy over the \InfoNCE baseline on all four benchmarks, despite the two methods sharing the same low-temperature regime. As \rmix increases, zero-shot accuracy degrades monotonically, consistent with the mAP and retrieval trends, indicating that larger mixing ratios over-regularize the embedding space, smoothing out the class-discriminative geometry alongside the modality gap. Notably, this trade-off is qualitatively different from increasing $\tau$: high-temperature \InfoNCE also reduces the modality gap but devastates zero-shot performance (ImageNet Top-1 $51.9{\to}15.5\%$). Regularized \InfoNCE (Reg) similarly fails to improve over the baseline, suggesting that explicit alignment terms penalize the geometry in ways that are detrimental to transfer. Mixing, by contrast, achieves gap reduction through a change in the negative set rather than the loss landscape, preserving the low-temperature geometry that benefits zero-shot transfer while removing the degenerative divergence objective of \InfoNCE.

\paragraph{Practical implications.}
For applications prioritizing strict top‑1 retrieval (\eg, high‑precision search), our results suggest small mixing (\rmix $\approx$ 0.01–0.05); this configuration preserves CLIP‑level recall while reducing the modality gap. For tasks that require modality‑invariant representations or stronger semantic organization (\eg, linear probing, transfer, etc.), larger \rmix (\eg, 0.5) offer further gap reduction and higher class mAP at the cost of modest top‑rank recall. Finally, when tuning $\tau$  in the original \InfoNCE formulation we find that high temperatures can mimic the geometric effects of mixing but result in a steeper retrieval penalty.

\paragraph{Future work.}
Promising directions include (i) training a vision-language model using \XinfoNCE from scratch using larger batch sizes, as well as applying hard negatives from both modalities to improve discriminative performance \citep{FARTASH2018}. In addition, it would be interesting to explore the shared semantic space of both image and text, \eg by semantic editing. In CLIP, \citet{RADFORD_2021} state that no performance gains were made by adding a contrastive head to the CLIP model. Whether this is also true for \XinfoNCE remains to be seen.

\section{Conclusion}
We introduced \XinfoNCE, a simple modification of \InfoNCE that mixes intramodal and intermodal negatives, and analyzed how temperature and mixing jointly shape similarity distributions and, consequently, representation geometry. Empirically, \XinfoNCE consistently reduces the modality gap across temperatures and mixing ratios, while improving within-modality semanticity. For I2T and T2I retrieval in COCO, small mixing nearly preserves \InfoNCE retrieval at R@5/10 and incurs only a small R@1 drop; adding a light projection head restores and even improves top‑rank recall. Compared to increasing \InfoNCE temperature, mixing achieves a more favorable balance between gap reduction, linear‑probe quality, and retrieval.



\section*{Acknowledgements}
This work was supported by the Danish Data Science Academy, which is funded by the Novo Nordisk Foundation (NNF21SA0069429) and VILLUM FONDEN (40516). The research was futher supported by the Novo Nordisk Foundation grant NF22OC0076907 ”Cognitive spaces - Next generation explainability” and the Pioneer Centre for AI, DNRF grant number P1. Computational resources were provided by the DeiC National HPC (DeiC-DTU-N5-2024049).

\bibliography{main}

@String(CVPR  = {IEEE Conf. Comput. Vis. Pattern Recog.})

@String(ICLR  = {Int. Conf. Learn. Represent.})

@article{LIANG_a,
	title = {Mind the Gap: Understanding the Modality Gap in Multi-modal Contrastive Representation Learning},
	author = {Liang, Weixin and Zhang, Yuhui and Kwon, Yongchan and Yeung, Serena and Zou, James},
	langid = {english},
}

@inproceedings{WANG_2020,
  title={Understanding contrastive representation learning through alignment and uniformity on the hypersphere},
  author={Wang, Tongzhou and Isola, Phillip},
  booktitle={International conference on machine learning},
  pages={9929--9939},
  year={2020},
  organization={PMLR}
}

@inproceedings{
    SHI_2023,
    title={Towards understanding the modality gap in {CLIP}},
    author={Peiyang Shi and Michael C. Welle and M{\r{a}}rten Bj{\"o}rkman and Danica Kragic},
    booktitle={ICLR 2023 Workshop on Multimodal Representation Learning: Perks and Pitfalls},
    year={2023},
    url={https://openreview.net/forum?id=8W3KGzw7fNI}
}

@inproceedings{RADFORD_2021,
	title = {Learning Transferable Visual Models From Natural Language Supervision},
	issn = {2640-3498},
	url = {https://proceedings.mlr.press/v139/radford21a.html},
	eventtitle = {International Conference on Machine Learning},
	pages = {8748--8763},
	booktitle = {Proceedings of the 38th International Conference on Machine Learning},
	publisher = {{PMLR}},
	author = {Radford, Alec and Kim, Jong Wook and Hallacy, Chris and Ramesh, Aditya and Goh, Gabriel and Agarwal, Sandhini and Sastry, Girish and Askell, Amanda and Mishkin, Pamela and Clark, Jack and Krueger, Gretchen and Sutskever, Ilya},
	urldate = {2026-01-29},
	date = {2021-07-01},
	langid = {english},
}

@article{HUANG_,
	title = {Mind the Gap: Preserving and Compensating for the Modality Gap in {CLIP}-Based Continual Learning},
	author = {Huang, Linlan and Cao, Xusheng and Lu, Haori and Meng, Yifan and Yang, Fei and Liu, Xialei},
	langid = {english},
}

@misc{ROLE_2025,
	title = {Fill the Gap: Quantifying and Reducing the Modality Gap in Image-Text Representation Learning},
	url = {http://arxiv.org/abs/2505.03703},
	doi = {10.48550/arXiv.2505.03703},
	shorttitle = {Fill the Gap},
	number = {{arXiv}:2505.03703},
	publisher = {{arXiv}},
	author = {Role, François and Meyer, Sébastien and Amblard, Victor},
	urldate = {2026-01-29},
	date = {2025-05-06},
	eprinttype = {arxiv},
	eprint = {2505.03703 [cs]},
}

@misc{FAHIM_2024d,
	title = {It's Not a Modality Gap: Characterizing and Addressing the Contrastive Gap},
	url = {http://arxiv.org/abs/2405.18570},
	doi = {10.48550/arXiv.2405.18570},
	shorttitle = {It's Not a Modality Gap},
	number = {{arXiv}:2405.18570},
	publisher = {{arXiv}},
	author = {Fahim, Abrar and Murphy, Alex and Fyshe, Alona},
	urldate = {2026-01-30},
	date = {2024-06-06},
	eprinttype = {arxiv},
	eprint = {2405.18570 [cs]},
}

@article{ESLAMI_2024,
  title={Mitigate the gap: Investigating approaches for improving cross-modal alignment in clip},
  author={Eslami, Sedigheh and de Melo, Gerard},
  journal={arXiv preprint arXiv:2406.17639},
  year={2024}
}

@misc{LEVI_2025,
	title = {The Double-Ellipsoid Geometry of {CLIP}},
	url = {http://arxiv.org/abs/2411.14517},
	doi = {10.48550/arXiv.2411.14517},
	number = {{arXiv}:2411.14517},
	publisher = {{arXiv}},
	author = {Levi, Meir Yossef and Gilboa, Guy},
	urldate = {2026-01-30},
	date = {2025-05-24},
	eprinttype = {arxiv},
	eprint = {2411.14517 [cs]},
}

@inproceedings{WANG_2021a,
	location = {Nashville, {TN}, {USA}},
	title = {Understanding the Behaviour of Contrastive Loss},
	rights = {https://ieeexplore.ieee.org/Xplorehelp/downloads/license-information/{IEEE}.html},
	isbn = {978-1-6654-4509-2},
	url = {https://ieeexplore.ieee.org/document/9577669/},
	doi = {10.1109/CVPR46437.2021.00252},
	eventtitle = {2021 {IEEE}/{CVF} Conference on Computer Vision and Pattern Recognition ({CVPR})},
	pages = {2495--2504},
	booktitle = {2021 {IEEE}/{CVF} Conference on Computer Vision and Pattern Recognition ({CVPR})},
	publisher = {{IEEE}},
	author = {Wang, Feng and Liu, Huaping},
	urldate = {2026-01-30},
	date = {2021-06},
	langid = {english},
}

@article{MANNA_2025,
  title={Dynamically scaled temperature in self-supervised contrastive learning},
  author={Manna, Siladittya and Chattopadhyay, Soumitri and Dey, Rakesh and Pal, Umapada and Bhattacharya, Saumik},
  journal={IEEE Transactions on Artificial Intelligence},
  volume={6},
  number={6},
  pages={1502--1512},
  year={2025},
  publisher={IEEE}
}

@inproceedings{WU_2018,
	location = {Salt Lake City, {UT}},
	title = {Unsupervised Feature Learning via Non-parametric Instance Discrimination},
	isbn = {978-1-5386-6420-9},
	url = {https://ieeexplore.ieee.org/document/8578491/},
	doi = {10.1109/CVPR.2018.00393},
	eventtitle = {2018 {IEEE}/{CVF} Conference on Computer Vision and Pattern Recognition ({CVPR})},
	pages = {3733--3742},
	booktitle = {2018 {IEEE}/{CVF} Conference on Computer Vision and Pattern Recognition},
	publisher = {{IEEE}},
	author = {Wu, Zhirong and Xiong, Yuanjun and Yu, Stella X. and Lin, Dahua},
	urldate = {2026-01-30},
	date = {2018-06},
	langid = {english},
}

@misc{OORD_2019,
	title = {Representation Learning with Contrastive Predictive Coding},
	url = {http://arxiv.org/abs/1807.03748},
	doi = {10.48550/arXiv.1807.03748},
	number = {{arXiv}:1807.03748},
	publisher = {{arXiv}},
	author = {Oord, Aaron van den and Li, Yazhe and Vinyals, Oriol},
	urldate = {2026-01-30},
	date = {2019-01-22},
	eprinttype = {arxiv},
	eprint = {1807.03748 [cs]},
}

@inproceedings{BACHMAN_2019,
	title = {Learning Representations by Maximizing Mutual Information Across Views},
	volume = {32},
	url = {https://proceedings.neurips.cc/paper_files/paper/2019/hash/ddf354219aac374f1d40b7e760ee5bb7-Abstract.html},
	booktitle = {Advances in Neural Information Processing Systems},
	publisher = {Curran Associates, Inc.},
	author = {Bachman, Philip and Hjelm, R Devon and Buchwalter, William},
	urldate = {2026-01-30},
	date = {2019},
}

@misc{CHEN_2020b,
	title = {A Simple Framework for Contrastive Learning of Visual Representations},
	url = {http://arxiv.org/abs/2002.05709},
	doi = {10.48550/arXiv.2002.05709},
	number = {{arXiv}:2002.05709},
	publisher = {{arXiv}},
	author = {Chen, Ting and Kornblith, Simon and Norouzi, Mohammad and Hinton, Geoffrey},
	urldate = {2026-01-30},
	date = {2020-07-01},
	eprinttype = {arxiv},
	eprint = {2002.05709 [cs]},
}

@article{HE_,
  title = {Momentum {{Contrast}} for {{Unsupervised Visual Representation Learning}}},
  author = {He, Kaiming and Fan, Haoqi and Wu, Yuxin and Xie, Saining and Girshick, Ross},
  langid = {english}
}

@misc{EMNIST,
	title = {{EMNIST}: an extension of {MNIST} to handwritten letters},
	url = {http://arxiv.org/abs/1702.05373},
	doi = {10.48550/arXiv.1702.05373},
	shorttitle = {{EMNIST}},
	number = {{arXiv}:1702.05373},
	publisher = {{arXiv}},
	author = {Cohen, Gregory and Afshar, Saeed and Tapson, Jonathan and Schaik, André van},
	urldate = {2026-02-07},
	date = {2017-02-17},
	eprinttype = {arxiv},
	eprint = {1702.05373 [cs]},
	note = {version: 1},
}

@misc{MSCOCO,
	title = {Microsoft {COCO}: Common Objects in Context},
	url = {http://arxiv.org/abs/1405.0312},
	doi = {10.48550/arXiv.1405.0312},
	shorttitle = {Microsoft {COCO}},
	number = {{arXiv}:1405.0312},
	publisher = {{arXiv}},
	author = {Lin, Tsung-Yi and Maire, Michael and Belongie, Serge and Bourdev, Lubomir and Girshick, Ross and Hays, James and Perona, Pietro and Ramanan, Deva and Zitnick, C. Lawrence and Dollár, Piotr},
	urldate = {2026-02-07},
	date = {2015-02-21},
	eprinttype = {arxiv},
	eprint = {1405.0312 [cs]},
}

@online{OH_2023,
  title = {Geodesic {{Multi-Modal Mixup}} for {{Robust Fine-Tuning}}},
  author = {Oh, Changdae and So, Junhyuk and Byun, Hoyoon and Lim, YongTaek and Shin, Minchul and Jeon, Jong-June and Song, Kyungwoo},
  date = {2023-11-07},
  eprint = {2203.03897},
  eprinttype = {arXiv},
  eprintclass = {cs},
  doi = {10.48550/arXiv.2203.03897},
  url = {http://arxiv.org/abs/2203.03897},
  urldate = {2026-02-12},
  pubstate = {prepublished}
}

@inproceedings{GRASSUCCI_2025,
  title={Closing the modality gap enables novel multimodal learning applications},
  author={Grassucci, Eleonora and Cicchetti, Giordano and Comminiello, Danilo},
  booktitle={Second Workshop on Representational Alignment at ICLR 2025},
  year={2025}
}

@software{CLIP_GH,
  title = {Openai/{{CLIP}}},
  date = {2026-02-07T14:39:48Z},
  origdate = {2020-12-16T11:24:42Z},
  url = {https://github.com/openai/CLIP},
  urldate = {2026-02-07},
  organization = {OpenAI}
}

@misc{UMAP,
      title={UMAP: Uniform Manifold Approximation and Projection for Dimension Reduction},
      author={Leland McInnes and John Healy and James Melville},
      year={2020},
      eprint={1802.03426},
      archivePrefix={arXiv},
      primaryClass={stat.ML},
      url={https://arxiv.org/abs/1802.03426},
}

@misc{FARTASH2018,
      title={VSE++: Improving Visual-Semantic Embeddings with Hard Negatives},
      author={Fartash Faghri and David J. Fleet and Jamie Ryan Kiros and Sanja Fidler},
      year={2018},
      eprint={1707.05612},
      archivePrefix={arXiv},
      primaryClass={cs.LG},
      url={https://arxiv.org/abs/1707.05612},
}

@misc{KAPARTHY,
      title={Deep Visual-Semantic Alignments for Generating Image Descriptions},
      author={Andrej Karpathy and Li Fei-Fei},
      year={2015},
      eprint={1412.2306},
      archivePrefix={arXiv},
      primaryClass={cs.CV},
      url={https://arxiv.org/abs/1412.2306},
}

@software{DELPHBOY,
  title = {Delphboy/Karpathy-Splits},
  author = {Senior, Henry},
  date = {2026-01-10T04:23:53Z},
  origdate = {2023-09-18T11:37:32Z},
  url = {https://github.com/Delphboy/karpathy-splits},
  urldate = {2026-02-22}
}

@article{DOSOVITSKY_2020,
  title={An image is worth 16x16 words: Transformers for image recognition at scale},
  author={Dosovitskiy, Alexey and Beyer, Lucas and Kolesnikov, Alexander and Weissenborn, Dirk and Zhai, Xiaohua and Unterthiner, Thomas and Dehghani, Mostafa and Minderer, Matthias and Heigold, Georg and Gelly, Sylvain and others},
  journal={arXiv preprint arXiv:2010.11929},
  year={2020}
}
\bibliographystyle{tmlr}

\clearpage
\appendix

\section{Hyperparameters}
\label{sec:appendix_hyperparams}
Key hyperparameters for both Image-to-Image and Image-to-Text experiments.

\setlength{\tabcolsep}{4pt}
\begin{table}[H]
	\centering
	\caption{Essential hyperparameters for Image-to-Image (\cref{sec:mnist_experiments}) and Image-to-Text (\cref{sec:coco_experiments}) experiments.}
	\label{tab:hyperparams}
	\begin{subtable}{0.5\linewidth}
		\centering
		\caption{Model parameters.}
		\label{tab:model_params}
		\begin{tabular}{lrr}
			\toprule
			            & EMNIST & COCO     \\
			\midrule
			Model name  & ViT    & ViT-B/16 \\
			Num. param. & 116 K  & 149 M    \\
			Patch size  & 4      & 16       \\
			Emb. dim.   & 48     & 512      \\
			Num. heads  & 4      & 12       \\
			Layers      & 4      & 12       \\
			MLP ratio   & 4      & 4        \\
			\bottomrule
		\end{tabular}
	\end{subtable}
	\begin{subtable}{0.49\linewidth}
		\centering
		\caption{Optimization parameters.}
		\label{tab:opt_params}
		\begin{tabular}{lrr}
			\toprule
			              & EMNIST & COCO  \\
			\midrule
			Optimizer     & AdamW  & AdamW \\
			Batch size    & 2048   & 576   \\
			Num epochs    & 250    & 30    \\
			\;\;\;Phase 1 & 1      & 1     \\
			\;\;\;Phase 2 & 5      & 1     \\
			\;\;\;Phase 3 & 244    & 28    \\
			Max. lr.      & 5e-4   & 2e-5  \\
			\bottomrule
		\end{tabular}
	\end{subtable}
\end{table}

\section{Regularized InfoNCE}
\label{sec:appendix_regularization}

The regularized \InfoNCE (Reg) baseline developed by \citet{FAHIM_2024d} augments the standard symmetric \InfoNCE objective with three penalty terms inspired by \citet{WANG_2020}:
\begin{equation}
    \mathcal{L}_{\text{Reg}} = \mathcal{L}_{\text{InfoNCE}} + \lambda_u\,\mathcal{L}_{\text{unif}} + \lambda_{xu}\,\mathcal{L}_{x\text{-unif}} + \lambda_a\,\mathcal{L}_{\text{align}},
\end{equation}
where all weights are set to $\lambda_u{=}\lambda_{xu}{=}\lambda_a{=}1$ in our experiments.
The three terms are defined as follows.

\paragraph{Uniformity} $\mathcal{L}_{\text{unif}}$ measures how uniformly each modality's embeddings are distributed on the unit hypersphere. Following \citet{WANG_2020}, it is defined as the log of the expected pairwise Gaussian potential within each modality:
\begin{equation}
    \mathcal{L}_{\text{unif}} = \frac{1}{2}\log\,\mathbb{E}\!\left[e^{-2\|\u_i - \u_j\|_2^2}\right] + \frac{1}{2}\log\,\mathbb{E}\!\left[e^{-2\|\v_i - \v_j\|_2^2}\right],
\end{equation}
where expectations are over distinct pairs $i \neq j$. This loss is minimised when embeddings within each modality are spread uniformly over the hypersphere.

\paragraph{Cross-modal uniformity} $\mathcal{L}_{x\text{-unif}}$ extends the uniformity criterion across modalities, penalising systematic clustering of image embeddings near text embeddings (or vice versa):
\begin{equation}
    \mathcal{L}_{x\text{-unif}} = \log\,\mathbb{E}\!\left[e^{-2\|\u_i - \v_j\|_2^2}\right],
\end{equation}
where the expectation is over all cross-modal pairs $(i, j)$ including $i{=}j$. This term directly discourages the collapse of inter-modal distances, acting as an explicit counterforce to the modality gap.

\paragraph{Alignment} $\mathcal{L}_{\text{align}}$ is the expected $\ell_2$ distance between $\ell_2$-normalised positive pairs $(\u_i, \v_i)$:
\begin{equation}
    \mathcal{L}_{\text{align}} = \mathbb{E}\left[\|\u_i - \v_i\|_2\right].
\end{equation}

\section{Convergence}
\label{sec:appendix_convergence}

We show the modality distance for \InfoNCE, regularized \InfoNCE (Reg), and \XinfoNCE (Mix) during training. While the modality distance drops almost instantly for models trained using \XinfoNCE (Mix), the modality distance decreases more slowly for regularized \InfoNCE, as the regularization term counteracts the divergent properties of \InfoNCE rather than fixing it directly.

\begin{figure}[H]
	\centering
	\includegraphics[width=0.8\textwidth]{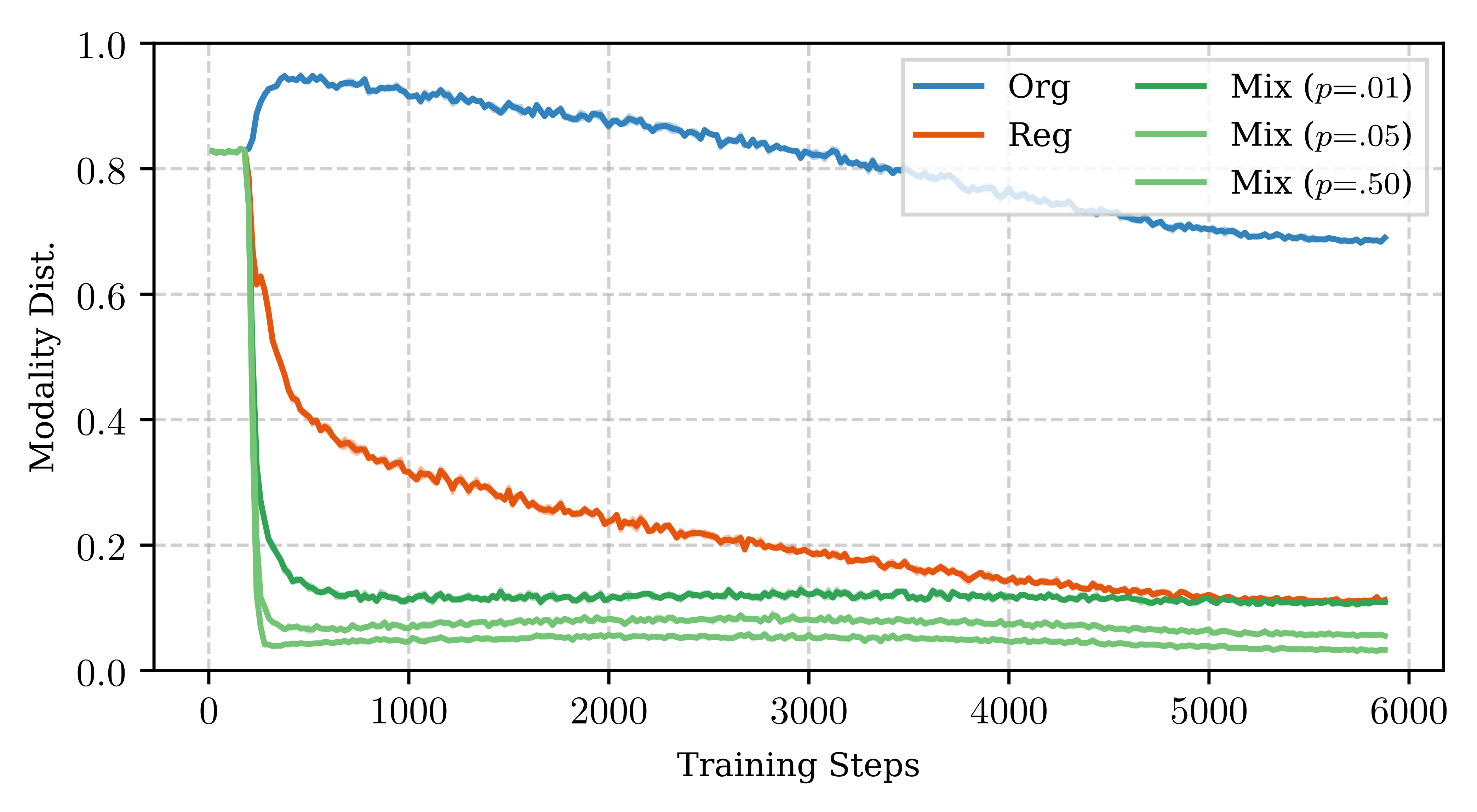}
	\caption{Modality distance progression during training. All models are initialized using pretrained CLIP weights. No weight updates for the first 200 steps.}
	\label{fig:modality_dist_progression}
\end{figure}

\section{Sensitivity Analysis}
\label{sec:sensitivity}

We observe a trade-off between retrieval performance and modality distance for various mixing ratios. Higher mixing ratios lead to lower modality distance but worse retrieval performance, and vice versa. However, even for the smallest mixing ratio, the modality distance is significantly smaller than for \InfoNCE (0.06 for \XinfoNCE vs. 0.88 for \InfoNCE).

\begin{figure}[H]
	\centering
	\includegraphics[width=0.8\textwidth]{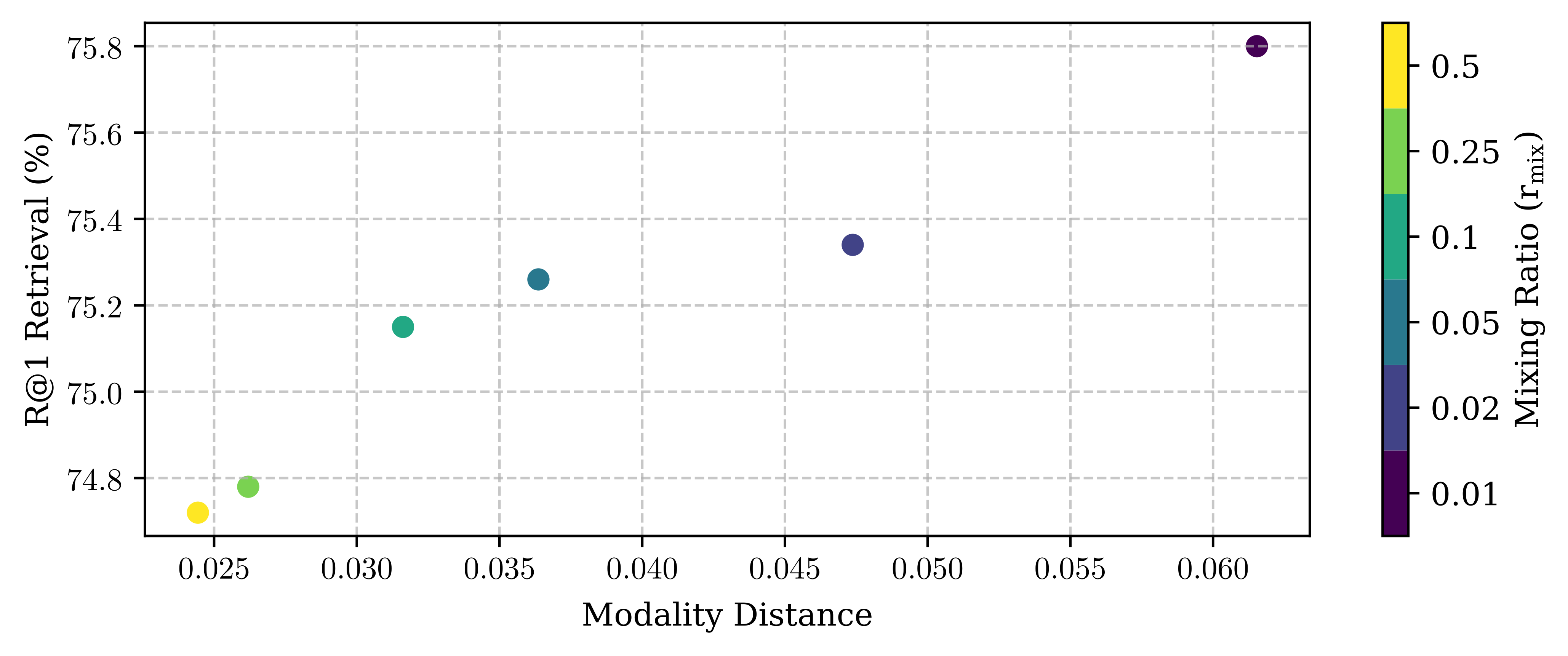}
	\caption{Sensitivity of retrieval performance to \rmix. Retrieval recall R@1 vs. modality distance for various mixing ratios. There is a clear trade-off between retrieval performance and modality distance.}
	\label{fig:sensitivity}
\end{figure}

\end{document}